\documentclass[acmtog, nonacm=true]{acmart}
\acmSubmissionID{345}

\usepackage{booktabs} 
\usepackage{multirow}
\usepackage{enumitem}
\usepackage{wrapfig,blindtext} 
\usepackage{hyperref}
\hypersetup{
    colorlinks=true,
    linkcolor=blue,   
    urlcolor=blue,
}
\definecolor{mypink1}{rgb}{0.858, 0.188, 0.478}
\usepackage[ruled]{algorithm2e} % For algorithms
\usepackage{makecell, cellspace, caption}

\SetAlFnt{\small}
\SetAlCapFnt{\small}
\SetAlCapNameFnt{\small}
\SetAlCapHSkip{0pt}

\acmJournal{TOG}
\acmVolume{40}
\acmNumber{4}
\acmArticle{83}
\acmYear{2021}
\acmMonth{8}

\setcopyright{rightsretained} 

\acmDOI{10.1145/3450626.3459825}

\begin{document}  
\title{Neural Monocular 3D Human Motion Capture with Physical Awareness}
 
\author{Soshi Shimada}
\affiliation{%
 \institution{Max Planck Institute for Informatics, Saarland Informatics Campus}
   \city{Saarbrücken} 
  \country{Germany}
 }
 \email{sshimada@mpi-inf.mpg.de}
 \author{Vladislav Golyanik}
\affiliation{%
 \institution{Max Planck Institute for Informatics, Saarland Informatics Campus}
  \city{Saarbrücken} 
  \country{Germany}
 }
 \email{golyanik@mpi-inf.mpg.de}
 \author{Weipeng Xu}
\affiliation{%
 \institution{Facebook Reality Labs}
   \city{Pittsburgh} 
  \country{USA}
 }
 \email{xuweipeng@fb.com}
  \author{Patrick Pérez}
\affiliation{%
 \institution{Valeo.ai}
    \city{Paris} 
  \country{France}
 } 
 \email{patrick.perez@valeo.com}
 \author{Christian Theobalt}
\affiliation{%
 \institution{Max Planck Institute for Informatics, Saarland Informatics Campus}
   \city{Saarbrücken} 
  \country{Germany}
 } 
 \email{theobalt@mpi-inf.mpg.de} 
 \begin{teaserfigure}
\centering 
\includegraphics[width=\linewidth]{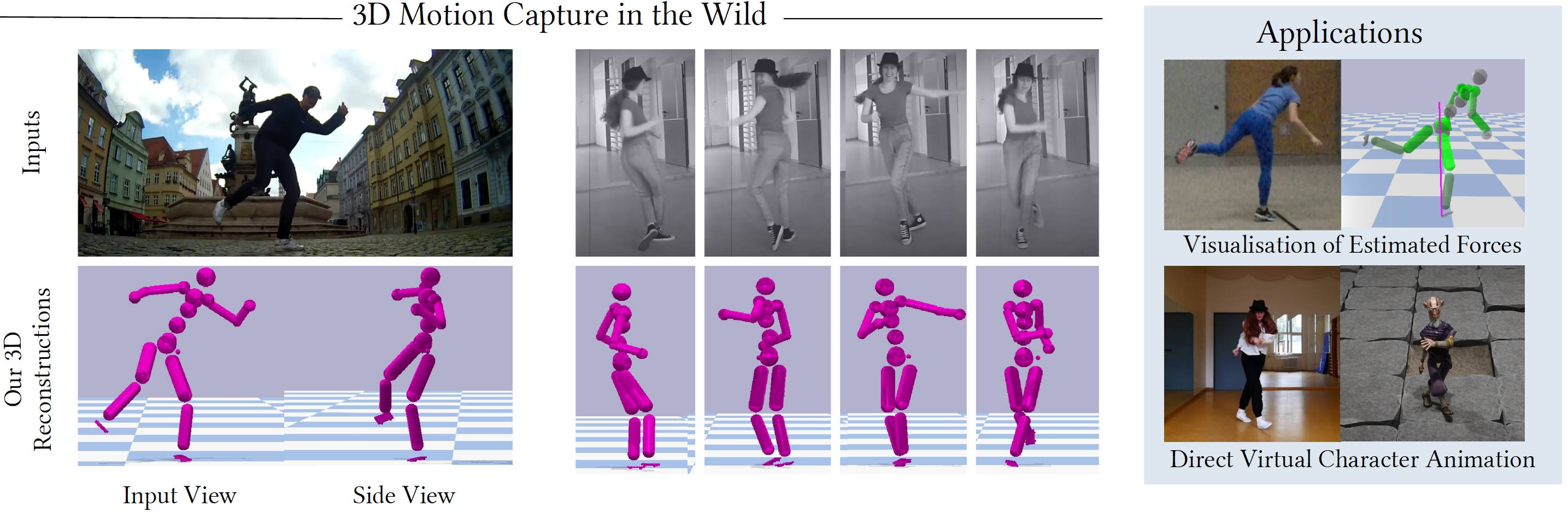} 
\caption{\label{fig:teaser} From an input monocular video, our method for markerless 3D human motion capture estimates global human poses which obey (bio-)physical constraints. 
In contrast to existing methods with physical awareness, our approach is neural and fully differentiable; it allows learning motion priors and the associated physical properties from the data. 
We can reconstruct more challenging and faster motions compared to the state of the art, with fewer artefacts such as jitter, foot-floor penetration and unnatural body postures. 
Thanks to these properties, our method can be used to directly drive a virtual character or visualise joint torques. 
(Left:) Results of our method on different sequences from the input and side views. 
(Right:) Applications in motion analysis by force visualisation and virtual character animation. 
} 
\label{fig:physcap_comparison} 
\end{teaserfigure} 

\begin{abstract} 
We present a new trainable system for physically plausible markerless 3D human motion capture, which achieves state-of-the-art results in a broad range of challenging scenarios. 
Unlike most neural methods for human motion capture, our approach, which we dub “physionical”, is aware of physical and environmental constraints. 
It combines in a fully-differentiable way several key innovations, \textit{i.e.}, 1) a proportional-derivative controller, with gains predicted by a neural network, that reduces delays even in the presence of fast motions, 2) an explicit rigid body dynamics model and 3) a novel optimisation layer that prevents physically implausible foot-floor penetration as a hard constraint. 
The inputs to our system are 2D joint keypoints, which are canonicalised in a novel way so as to reduce the dependency on intrinsic camera parameters---both at train and test time. 
This enables more accurate global translation estimation without generalisability loss. 
Our model can be finetuned only with 2D annotations when the 3D annotations are not available. 
It produces smooth and physically-principled 3D motions in an interactive frame rate in a wide variety of challenging scenes, including newly recorded ones. Its advantages are especially noticeable on in-the-wild sequences that significantly differ from common 3D pose estimation benchmarks such as Human 3.6M and MPI-INF-3DHP.
Qualitative results are available at \textcolor{mypink1}{ \url{http://gvv.mpi-inf.mpg.de/projects/PhysAware/}}
\end{abstract}




\maketitle

\section{Introduction}\label{sec:intro}  
3D human motion capture is an actively researched area enabling many applications ranging from human activity recognition to sports analysis, virtual-character animation, film production, human-computer interaction and mixed reality.  
Since marker-based and multi-camera-based solutions are expensive and unsuited for many applications (\textit{e.g.}, in-the-wild capture and recordings outside the studio or legacy content), 
methods for \textit{markerless} 3D human motion capture from a monocular camera \cite{VNect_SIGGRAPH2017, shimada2020physcap} are intensively researched. 

Monocular 3D human motion capture is a highly challenging inverse problem, due to the fundamental ambiguities in deducing 3D body configuration and scale from 2D cues, as well as due to difficult (self-)occlusions \cite{VNect_SIGGRAPH2017, JMartinezICCV_2017, pavlakos2018learning, kocabas2019vibe}.  
Most state-of-the-art methods for 3D human pose estimation and motion capture benefit from the rapid progress in machine learning and have shown stark improvements in accuracy \cite{WandtRosenhahn2019, sun2019human, kocabas2019vibe}. 
Despite this progress, existing predominantly purely kinematic methods still have important limitations and produce notable artefacts.   
Many produce per-frame predictions that can be temporally highly unstable, and many produce root-relative but not global 3D poses.  
Further, most existing methods are incapable by design to consider interactions with the environment, let alone biophysical pose or motion plausibility. 
The former often leads to collision violations such as foot-floor penetration and floating in the air in captured motions, the latter yields impossible poses with physically-impossible leaning and posture, or poses that would actually cause loss of balance.  
Captured results are therefore not only inaccurate in several ways but also unnatural, which greatly reduces data usability, in particular in computer graphics related applications. 

We, therefore, propose a new neural network-based approach for monocular 3D  human motion capture which considers physical  constraints in the observed scenes, see  Fig.~\ref{fig:teaser} for an overview.  

We believe that improving upon the recently proposed ideas of physical-awareness constraints in monocular 3D human motion capture \cite{rempe2020contact, shimada2020physcap} and combining them with machine learning techniques can lead to further advances in the domain. 
While the methods of \cite{rempe2020contact, shimada2020physcap} contain two stages---with the physics-based pose optimisation implemented as an engineered method relying on classical optimisation techniques,---\textit{we are the first to propose a fully-differentiable framework for monocular 3D human motion capture with physical awareness}. 
Thus, our physics-based pose optimisation is a trainable neural network with custom layers for physics-based constraints. 
We refer to our approach as \textit{physionical}, which means that it is fully-differentiable, neural network-based and aware of physical boundary conditions. The 3D motions estimated by our framework are smooth and natural, and can directly drive an animation character with no further postprocessing. We can also visualise the joint torques and ground reaction forces estimated from the motion in the video, which can be used for some applications, \textit{e.g.}, sports analysis. See Fig. \ref{fig:teaser} for the visualisation of the reconstructed 3D motions and the example applications of our framework.  

Our method includes two core neural components, \textit{i.e.}, a  \textit{target pose estimator network} (TPNet) and an iterative \textit{dynamic cycle} for controlling a humanoid character while considering physics-based boundary conditions. 
Both TPNet and the dynamic cycle are newly developed neural networks that are end-to-end trained. 
TPNet kinematically regresses the target reference 3D pose from input 2D keypoints that are obtained by an off-the-shelf 2D detector, which serves as a foundation for the dynamic cycle. 
The dynamic cycle first calculates gain parameters of a neural  proportional-derivative (PD) controller which generates a force vector to control the kinematic character with physics properties through the differentiable physics model.
The force vector is then used to estimate the ground reaction force (GRF), and both 
are then passed to the forward dynamics module which regresses the accelerations of the skeleton. 
The latter are subsequently used to update the final global human pose in 3D which matches the subject's 2D pose in the input frames and obeys the condition of plausible foot-floor placements. 
In the dynamic cycle our architecture contains a novel differentiable layer realising a \textit{hard} constraint for preventing foot-floor penetration. 
Our motivation for a custom optimisation layer comes from the fact that conventional losses in neural networks can only express soft constraints on the learned manifold, \textit{i.e.}, there is no guarantee that the expressed boundary conditions will be strictly fulfilled at inference time. 
On the other hand, physical constraints and forces such as gravity and ground reaction force (originating from the floor which naturally limits human motions) are strictly present in the physical world without freedom of interpretation. 

Since our architecture is fully differentiable, it is the first approach for monocular physics-aware 3D motion capture that can be  equally trained on images annotated with strong and weak labels, \textit{i.e.}, 
joint angles, 3D joint keypoints but also 2D joint keypoints. 
Since also 2D training data can be used, our method can be trained for better generalisation and is easier to fine-tune for motion classes for which any 3D annotation would be very hard (\textit{e.g.}, in-the-wild athletics or sports videos). 
Our physionical method is aware of the environment and physical laws and runs in real time at $20$ frames per second. 
It outputs physically-plausible results with significantly fewer artefacts---such as unnatural temporal instabilities and frame-to-frame jitter, foot-floor penetration and the uncertainty in the absolute human poses along the depth dimension---than purely kinematic methods and other physics-aware methods. 
Moreover, compared to the previous most related method PhysCap  \cite{shimada2020physcap}, we mitigate the delay between the observed and estimated motions. 
To summarise, the technical \textbf{contributions} of this article are as follows: 
\begin{itemize}[leftmargin=15pt] 
    \item The first % 
    entirely-neural and fully-differentiable 
    approach for markerless 3D human motion capture from monocular videos with physics constraints, which we call physionical (Sec.~\ref{sec:Method}). 
    \item A new canonicalisation of input 2D keypoints allowing network training and 3D human pose regression with different intrinsic camera parameters and jointly on several datasets (Sec.~\ref{ssec:canonicalisation}). 
    In contrast to existing normalisation methods for human pose estimation in the literature, 
    our canonicalisation does not discard the cues for the global pose estimation. 
    \item The integration of hard boundary conditions in our architecture to prevent foot-floor penetrations 
    by taking advantage of the recent progress in designing optimisation layers for neural architectures \cite{cvxpylayers2019} (Sec.~\ref{ssec:DynCycle}). 
    \item Applications of our method in 
    direct virtual character animation and 
    visualisation of joint torques related to muscle activation forces, which can be used to analyse the captured motions in conceivable downstream tasks  (Sec.~\ref{ssec:applications}). 
\end{itemize} 
 
The proposed method establishes a new state of the art and outperforms existing methods on several metrics, as shown in our experiments (Sec.~\ref{sec:experiments}). 
We evaluate it on several datasets including Human3.6M \cite{ionescu2013human3}, MPI-INF-3DHP \cite{mono-3dhp2017}, DeepCap \cite{deepcap} as well as newly-recorded sequences (Sec.~\ref{sec:experiments}). 
The differences in the results of our physionical approach compared to existing techniques are especially noticeable when they are obtained on scenes in the wild. 
See our supplementary video with visualisations of the experimental results. 

\section{Related Work}\label{sec:Related_works} 
A vast body of literature is devoted to 3D human motion capture with multi-view systems  \cite{starck2007surface, Bo2008, brox2010combined, Gall2010, Wu2012, elhayek2015efficient,  Brualla2018} and inertial on-body sensors \cite{Dejnabadi2006, Vlasic2007, Tautges2011, vonMarcard2017}. 
Both areas are well studied and these methods have shown impressive results. 
On the downside, they require specialised camera rigs and hardware which make their operation  outside the studio difficult.  
In this section, we thus further focus on related works on 1) physics-based virtual character  animation and 2) monocular 3D human pose estimation and motion capture. 

\paragraph{Physics-Based Virtual Character Animation} 
Many works have been proposed for physics-based character animation which is a  significantly different problem compared to monocular 3D human motion capture. 
In virtual character animation, there is full control over the simulated physical laws and the structure of the simulated world (in which virtual characters are moving), whereas we are interested in reconstructing physically-plausible human motions from partial observations (monocular videos). 
At the same time, the animated character of these methods is inspirational for us, as they provide the realism and motion  plausibility of character motion required in computer graphics applications \cite{Barzel1996, Sharon2005Walking, Wrotek2006, Liu2010Samcon, levine2012physically, zheng2013human, andrews2016real, bergamin2019drecon}. 
Some techniques for virtual character animation employ deep  reinforcement learning and motion imitation in physics engines,  often requiring specialised networks for each motion kind 
\cite{Peng2018,peng2018deepmimic,bergamin2019drecon,Lee2019,Jiang2019}. 
In contrast to the latter, our problem requires a different approach. 
Since our goals are the generalisability across different motions and high data throughout enabling real-time applications, we use explicit equations of motions and physics-based constraints on top of initial kinematic estimates, while preserving the differentiability of our architecture trained in a supervised manner. 

\paragraph{Classical Monocular 3D Human Motion Capture and Pose Estimation} 
This section focuses on the majority of works on monocular 3D human motion capture and pose estimation that do not use explicit physics-based and environment constraints. 
All such methods for 3D human pose estimation and motion capture can be classified into 1)  direct regression approaches, 2) lifting approaches and 3) various hybrid approaches leveraging mixtures of 3D and 2D predictions. 
The first category of methods is based on convolutional neural networks and 
regresses 3D joints directly from input images \cite{tekin2016structured, mono-3dhp2017, rhodin2018unsupervised}. 
The methods of the second category regress 3D joints from detected 2D keypoints  \cite{chen_2017_3d, JMartinezICCV_2017, tome2017lifting, Moreno-Noguer_2017, pavlakos2018learning}. 
Finally, multiple methods combine 3D joint depth (or location probabilities) and 2D keypoint prediction with lifting constraints \cite{newell2016stacked,  VNect_SIGGRAPH2017, pavlakos2017volumetric, Yang_3dposeCVPR2018, Zhou_2017_ICCV, inthewild3d_2019}. 
Among them, VNect \cite{VNect_SIGGRAPH2017, inthewild3d_2019} uses  additional weak supervision with in-the-wild images. 
Some methods additionally use 3D shape priors. 
Statistical human body models provide strong constraints on plausible human postures which can be used for human pose estimation \cite{bogo2016smpl, hmrKanazawa17, kocabas2019vibe}. 
\cite{Xu2018MHP, deepcap, EventCap2020} leverage actor-specific 3D human body templates for global 3D human motion capture with shape tracking including surface deformations on top of a skeletal motion. 
Several further algorithms use different variants of anatomical constraints for the human body (\textit{e.g.}, body symmetry) and show improved results in weakly-supervised \cite{Dabral:ECCV:2018, WandtRosenhahn2019} or even unsupervised 3D human pose estimation \cite{Kovalenko2019}. 
\cite{zhang2020phosa, PROX:2019} use geometric vicinity and collision avoidance constraints for the reconstruction of human-object interactions, and \cite{Zanfir2018, RogezWS18, Dabral2019, XNect_SIGGRAPH2020, Fabbri2020} can generalise to multiple persons in the scene. 
Most of the proposed algorithms work on single images  \cite{kolotouros2019spin,hmrKanazawa17,pavlakos2018learning,sun2019human,song2020human}, whereas others  take the temporal information into account for improved temporal stability  \cite{humanMotionKanazawa19,kocabas2020vibe,pavllo20193d}. 
To directly drive a kinematic character with skinned rigs, we need joint angles, root translation and rotation of a consistent skeleton. 
Only few works estimate those from the input RGB video directly and realise the character motion control from a video  \cite{VNect_SIGGRAPH2017,XNect_SIGGRAPH2020,shi2020motionet}. 
Among the latter, MotioNet of Shi \textit{et  al.}~\shortcite{shi2020motionet} is the most closely  related method to ours. 
Unlike our approach, it does not include an explicit physics model, which adds up to physically-implausible effects in the estimates. 
Upon the architecture design, MotioNet expects at testing the same intrinsic camera parameters as in the training dataset, \textit{i.e.}, when the system is applied to sequences with different camera intrinsics, the accuracy of the estimated  translations can vary considerably. 
In contrast, we use canonical 2D keypoints 
which makes our physionical approach invariant to camera intrinsics. 

\paragraph{Monocular 3D Human Motion Capture with Physics-based Constraints.} 
This section focuses on the emerging field of monocular 3D human motion capture with physics-based constraints. 
One of the pioneering works in this domain 
was proposed by \citet{wei2010videomocap} back in 2010. 
Their method requires manual user interactions for each input sequence and is  computationally expensive. 
\citet{vondrak2012video} perform 3D human motion capture from monocular videos for physically-plausible character control. 
They recover 3D bipedal controllers using optimal control theory, which are capable of simulating the observed motions in different environments. 
Unfortunately, this method cannot easily generalise across motions and does not run in real time. 
\citet{ZelWan2017} estimate 3D human poses along with the inner and  exterior forces from images for object lifting and walking. 
\citet{li2019motionforcesfromvideo} regress human and object poses in 3D along with  forces and torques exerted by human limbs from a monocular video and an object prior. 
They focus on instruments with grips and recognise contacts between a person and an object (\textit{i.e.}, the instrument or the ground) to facilitate the trajectory-optimisation problem. 
The method of \citet{Zell2020} for the analysis of 3D human motion capture relates to our setting. 
It infers ground-reaction forces and joint torques from input 3D human motion capture sequences. 
It relies on a new dataset with multiple human motion types and ground-truth forces acquired using force plates on the floor. 
The advantage of this method is that the proposed forward and inverse dynamics layers generalise to new locomotion types. 
Thus, the main focus lies on the \textit{explainability} of the captured human motions in 3D from the physics perspective, whereas our goal is 3D human motion capture that satisfies physics-based (environmental) constraints at  interactive framerates. 
Two recent methods for monocular 3D human motion  capture with physics constraints are \cite{rempe2020contact} and \cite{shimada2020physcap}. 
They tackle general human motions by introducing laws of  physics as regularisers in their formulations. 
Both methods 1) start with initial kinematic estimates (\cite{Xiang2019} and \cite{VNect_SIGGRAPH2017}, respectively) which are subsequently refined through physics-based optimisation, 2) detect foot contacts and 3) assume that orientation of the ground plane is known (the final position can be refined), the camera is not moving and the entire human body is visible in all frames. 
\cite{rempe2020contact} and \cite{shimada2020physcap}, however, differ significantly in  physics-based global pose optimisation and the overall runtime. 
\citet{rempe2020contact} use as a proxy a reduced-dimensional model of the lower body inspired by \cite{Winkler2018}, which does not include all joints but captures the overall motion and contacts. 
In contrast, \citet{shimada2020physcap} rely on initial  kinematic pose corrections and a lightweight iterative physics-based pose refinement with PD joint controllers and ground-reaction force estimation, which enable real-time operation. 
Both these approaches are compositional and only partially rely on neural networks (for the kinematic estimates and foot contact detections, but not for the physics-based reasoning),  unlike our approach. 
We embed hard physics-based constraints through a custom layer in our architecture \cite{agrawal2019differentiable} and enable its full differentiability. 
Our trainable model with explicit physics-based constraints realises more plausible 3D motion qualitatively and more accurate 3D poses quantitatively than the existing physics-based approaches solving conventional optimisation problems with the dynamics equations of motion (see  Sec.~\ref{sec:experiments}). 

\section{Method}\label{sec:Method} 

\begin{figure*}[t!] 
\centering 
\includegraphics[width=\linewidth]{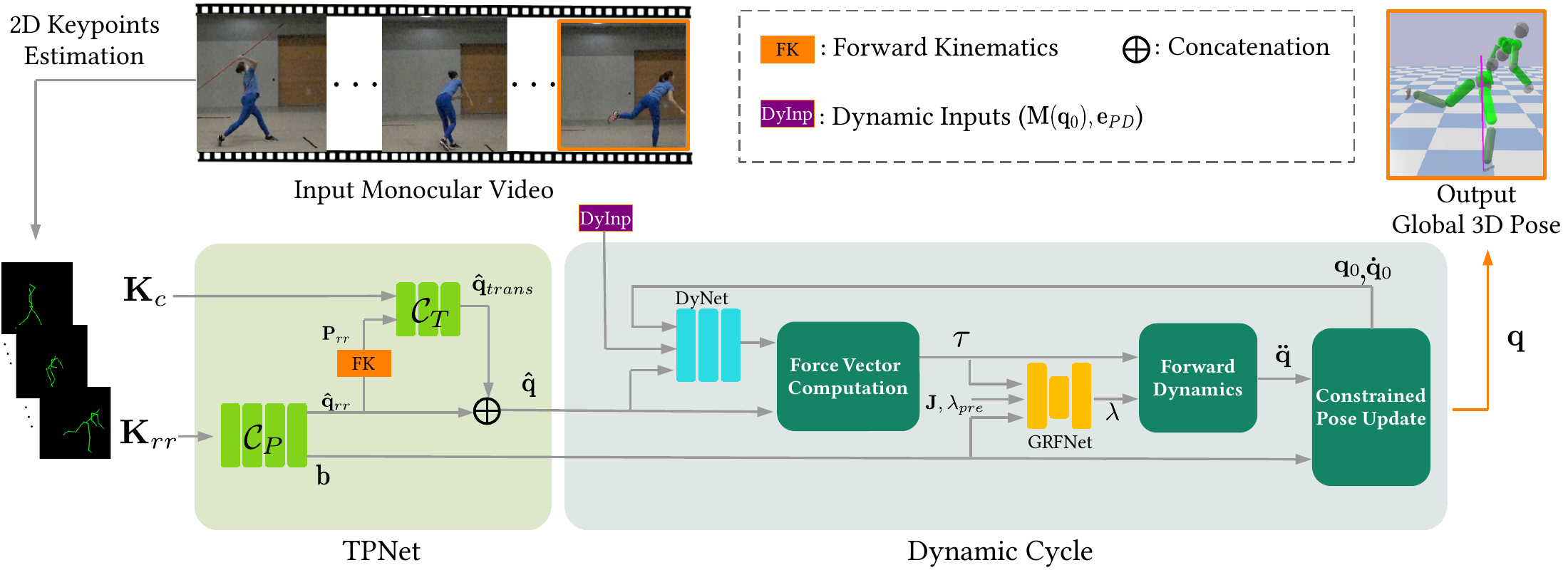} 
\caption{\textbf{Overview of our physionical approach for markerless monocular 3D human motion capture.}  
Our architecture assumes 2D keypoints in two representations as input, \textit{i.e.}, the canonicalised 2D keypoints ($\mathbf{K}_{c}$) and root-relative 2D  keypoints normalised by the image size ($\mathbf{K}_{rr}$). 
These representations are complementary and ensure that joint angles and root orientation can be accurately estimated (thanks to $\mathbf{K}_{rr}$) along with the global translation, with no dependency on the camera intrinsics (thanks to $\mathbf{K}_{c}$). 
First, the target kinematic pose $\mathbf{\hat{q}}$ is regressed with TPNet and fed to the dynamic cycle which implements various types of physics-based boundary conditions. 
The dynamic cycle includes several neural components. 
The input to DyNet is a set of parameters (the target pose $\mathbf{\hat{q}}$, the current pose $\mathbf{q}_{0}$, the current velocity $\dot{\mathbf{q}}_{0}$, the mass matrix $\mathbf{M}$ and the current pose error  $\mathbf{e}_{\text{PD}} = d(\mathbf{\hat{q}}, \mathbf{q}_{0})$) and the outputs are gain parameters $k_{p}$ of the PD controller and the offset force $\alpha$ for each DoF. 
The outputs from TPNet and DyNet are used to compute the force vector $\boldsymbol{\tau}$ following the PD controller rule. 
The GRFNet estimates the ground reaction force $\boldsymbol{\lambda}$. 
Both $\boldsymbol{\tau}$ and $\lambda$ are then 
passed to the forward dynamics module which regresses the accelerations $\ddot{\mathbf{q}}$ in the skeleton frame.  
This module considers mass matrix of the body $\mathbf{M}$, internal and external forces, gravity, Coriolis and centripetal forces. 
Finally, the character's pose is updated using the obtained $\ddot{\mathbf{q}}$ through the differentiable optimisation layer to prevent foot-floor penetration. 
} 
\label{fig:overview} 
\end{figure*}

\paragraph{Overview}
Our goal is physically-plausible monocular global 3D human motion capture without markers. 
We follow a learning-based approach trained through a fully-differentiable physics model, see 
Fig.~\ref{fig:overview} for an overview. 
Our framework includes a neural-network-based proportional-derivative (PD) controller that estimates a force vector allowing controlling the kinematic character with dynamics properties to match its pose with the subject's pose in the image sequence. 
The ground reaction forces are also estimated alongside the 3D motions without requiring any supervisory force annotations. 
We can also read out and visualise internal and contact forces regressed from the monocular input. 
Our method accepts sequential 2D joint keypoints in a video (\textit{e.g.}, extracted with an of-the-shelf 2D keypoint detector), and returns 3D skeleton poses which satisfy (bio-)physical constraints. 
This significantly mitigates foot-floor penetration, body sliding along depth direction and joint jitters. 
In Sec.~\ref{ssec:Model}, we define our model and mathematical  notations. 
In Sec.~\ref{ssec:canonicalisation}, we discuss a canonicalisation method of the input 2D joint keypoints which allows our global translation estimation network $\mathcal{C}_{T}$ to be trained jointly on several datasets 
with different camera intrinsics. 
In Secs.~\ref{sec:TPNet} and \ref{ssec:DynCycle}, the target pose estimation network and the dynamic cycle with physics-based constraints are elaborated, respectively. 
In the latter, the 3D pose is updated in the custom optimisation layer where we introduce a hard constraint to prevent foot-floor penetration in a differentiable manner. 
The obtained 3D poses are smooth, plausible and show mitigated motion delay even on fast motion sequences thanks to the learning-based PD controller which dynamically adjusts the gain parameters depending on the motions in the scene. 
Our fully-differentiable architecture allows finetuning using 2D  annotations only for improved accuracy on in-the-wild footage  (Sec.~\ref{ssec:in_the_wild}). 
Applications of our methods are discussed in  Sec.~\ref{ssec:applications}. 

\subsection{Our Model, Assumptions and Notations}\label{ssec:Model} 

We represent the kinematic state of the skeleton by a pose vector $\mathbf{q} \in \mathbb{R}^{n+1}$ and its velocity $\dot{\mathbf{q}} \in \mathbb{R}^{n}$ in the camera frame, with $n=46$.  
The first seven entries of $\mathbf{q}$ represent the root translation $\mathbf{q}_{\text{trans}} \in \mathbb{R}^3$ and rotation in the quaternion parametrisation $\mathbf{q}_{\text{ori}} \in  \mathbb{R}^{4}$, respectively.
All remaining $n - 7$ entries of $\mathbf{q}$ encode joint angles of the human skeleton model parametrised by Euler angles. 
The first three entries of $\dot{\mathbf{q}}$ represent the linear velocity of the root whereas the next three ones stand for its angular velocity $\mathbb{\omega} \in  \mathbb{R}^{3}$. 
The remaining entries of $\dot{\mathbf{q}}$ stand for the  angular velocity of each joint and they correspond to the joint order in $\mathbf{q}$. 
The time derivative of $\mathbf{q}_{\text{ori}}$ is approximated as follows: 
\begin{equation} \label{eq:ang_vel_quat} 
\frac{\mathrm{d}\mathbf{q}_{\text{ori}}}{\mathrm{d}\mathbf{t}}\approx  \frac{1}{2}\begin{bmatrix}
0\\ 
\omega
\end{bmatrix} \otimes   \, \mathbf{q}_{\text{ori}},
\end{equation} 
where $\otimes$ represents a quaternion multiplication.
Eq.~\eqref{eq:ang_vel_quat} is used to update the 3D root  orientation from its angular velocity in each dynamics simulation step. 

We use $M$ 2D joint keypoints normalised in two different ways, \textit{i.e.}, the root-relative 2D keypoints normalised by the image size and gathered in $\mathbf{K}_{rr} \in \mathbb{R}^{M\times 2}$, and the canonical 2D keypoints stacked in $\mathbf{K}_{c} \in \mathbb{R}^{M\times 2}$, 
allowing the network training on datasets with different camera intrinsics. 
Resorting to root-relative 2D joint keypoints is a widely-used normalisation approach for estimating the root-relative 3D pose from an image or video since it is translation-invariant in the image space. 
Therefore, we use $\mathbf{K}_{rr}$ for estimating the joint angles and root orientation of the character $\mathbf{q}_{rr}$. 
On the one hand, this normalisation alone loses the cues for estimating the global translation of the subject in the scene. 
On the other hand, canonicalised 2D joint keypoints retain the required information to regress the global pose, 
see Sec.~\ref{ssec:canonicalisation} for the details. 

Our character is composed of \textit{links} which are volumetric body part representations with collision proxies, following the same structure as \cite{shimada2020physcap}. 
Our core idea is to enable awareness of physical laws in our framework which helps to obtain physically-plausible human motion captures. 
We impose the laws of physics by considering Newtonian rigid body dynamics, which---when applied to our case---reads as \cite{featherstone2014rigid}: 
\begin{equation} \label{eq:eom}
   \mathbf{M}(\mathbf{q}) \ddot{\mathbf{q}} - \boldsymbol{\tau}   =   \mathbf{J}^{\mathsf{T}}(\mathbf{q}) \mathbf{G}\boldsymbol{\lambda} - \mathbf{h}(\mathbf{q},\dot{\mathbf{q}}),
\end{equation} 
where $\mathbf{M}\in \mathbb{R}^{n \times n}$ and $ \ddot{\mathbf{q}}\in \mathbb{R}^{n}$ are the inertia matrix in the skeleton frame, which describes the moments of inertia of the system, and the acceleration of $\mathbf{q}$, respectively; 
$\mathbf{J}\in \mathbb{R}^{6N_{c}\times n}$ is a contact Jacobi matrix which relates velocities in the skeleton frame to velocities in Cartisian coordinates; 
$N_{c}$ denotes the number of links to which the contact forces are applied; 
$\mathbf{G}\in \mathbb{R}^{6N_{c} \times 3N_{c}}$ is the matrix that converts the contact force  $\boldsymbol{\lambda}\in\mathbb{R}^{3N_{c}}$ to linear forces and torques 
(for its details, readers are referred to  \cite{featherstone2014rigid}); 
$\mathbf{h}\in \mathbb{R}^{n}$ encompasses gravity, Coriolis and centripetal forces; 
the force vector $\boldsymbol{\tau}\in \mathbb{R}^{n}$ represents the internal joint forces of the character, with its first six entries being the direct root actuations which are set to $0$ as per convention. 

The total forces that explain the root motion include  external forces such as ground reaction force (GRF). 
Similar to several prior works  \cite{shimada2020physcap,levine2012physically,andrews2016real,yuan2020residual}, we minimise the direct (virtual) root actuation by estimating the acting GRF and explaining the observed motions with it as much as possible (instead of setting the first six entries of $\boldsymbol{\tau}$ to zero). 

\subsection{Input Canonicalisation}\label{ssec:canonicalisation} 
For the networks that estimate the character's pose without global translation $\mathbf{q}_{rr}$ (\textit{e.g.}, $C_P$), we use root-relative 2D joint keypoints $\mathbf{K}_{rr}$.  
Many algorithms, which use a perspective camera model, estimate the global root position by optimising a 2D projection-based loss without learning components \cite{VNect_SIGGRAPH2017,deepcap,XNect_SIGGRAPH2020}. \cite{pavllo20193d} and \cite{shi2020motionet} employ neural networks to directly regress the translation of the 3D poses. 
However, in this case the learned motion manifolds 
depend on the camera intrinsic parameters used during the training. 
Consequently, at test time, they expect  similar camera intrinsics.
To tackle this issue, we propose to use canonicalised 2D keypoints $\mathbf{K}_{c}$ to factor out the influence of the camera intrinsics before they are fed to the neural network that regresses the absolute root translation of the character. 
Our architecture benefits from the canonicalisation in two ways. 
First, the translation estimation network can be trained with a large scale joint dataset, \textit{i.e.}, a composition of Human 3.6M \cite{ionescu2013human3}, MPI-INF-3D-HP \cite{mono-3dhp2017} and DeepCap \cite{deepcap}, which are recorded with different intrinsic camera parameters. 
Second, arbitrary camera intrinsics can be used at test time without influencing the performance of the network that regresses the global translations. 

Consider the perspective camera projection without a skew parameter: 
\begin{equation} \label{eq:normal_keys}  
 \begin{bmatrix}
f_{x} & 0 & c_{x} \\ 
 0 & f_{y} & c_{y}\\ 
 0 & 0  & 1
\end{bmatrix}\begin{bmatrix}
X\\ 
Y\\ 
Z
\end{bmatrix} = Z\begin{bmatrix}
\frac{f_{x}X}{Z}+c_{x}\\ 
\frac{f_{y}Y}{Z}+c_{y}\\ 
1
\end{bmatrix}, 
\end{equation} 
where $[X, Y, Z]^{\mathsf{T}}$ is a 3D coordinate of a joint in the camera frame,  
$f$ the focal length and $c$ the principal point. 
We see that the 2D joint keypoints $\Big[\frac{f_{x}X}{Z}+c_{x},  \frac{f_{y}Y}{Z}+c_{y}\Big]^{\mathsf{T}}$ are influenced by the camera intrinsic parameters.
Therefore, we generate canonical 2D joint keypoints by applying the identity as an intrinsic camera matrix: 
\begin{equation}\label{eq:canonical_keys}  
\begin{bmatrix}
1 & 0 & 0 \\ 
 0 & 1 & 0\\ 
 0 & 0  & 1
\end{bmatrix}\begin{bmatrix}
X\\ 
Y\\ 
Z
\end{bmatrix} = Z\begin{bmatrix}
\frac{X}{Z}\\ 
\frac{Y}{Z}\\ 
1
\end{bmatrix}. 
\end{equation}
We use $[X/Z, Y/Z]^{\mathsf{T}}$ as the canonical 2D joint keypoint which is not influenced by camera intrinsic parameters. 
In the case, when the depth information $Z$ is not known (\textit{e.g.}, during the testing phase), we can still obtain the canonical 2D joint keypoints assuming that camera intrinsics are known. 
Let $p_m = [u_m, v_m]^\mathsf{T}$ be the 2D joint locations of $M$ joints in the pixel coordinates, with $m \in \{1, \hdots, M\}$. 
We next stack the canonicalised 2D keypoints in a single $\mathbf{K}_{c}$ matrix:  
\begin{equation}\label{eq:canonical_keys_testing} 
    \mathbf{K}_{c} =  
        \Bigg[
        \begin{matrix}
            \frac{u_1-c_{x}}{f_{x}} \\
            \frac{v_1-c_{y}}{f_{y}}
        \end{matrix}  
        \;
        \begin{matrix}
            \frac{u_2-c_{x}}{f_{x}} \\
            \frac{v_2-c_{y}}{f_{y}}
        \end{matrix} 
        \;
        \hdots 
        \;
        \begin{matrix}
            \frac{u_M-c_{x}}{f_{x}} \\
            \frac{v_M-c_{y}}{f_{y}}
        \end{matrix}  
        \Bigg]^\mathsf{T}.  
\end{equation} 
It follows from Eqs.~\eqref{eq:canonical_keys} and \eqref{eq:canonical_keys_testing}, that for a single $p_m$ and for 
the corresponding 3D joint location $P_m = [X_m, Y_m, Z_m]^\mathsf{T}$, we have: 
\begin{equation}
        \begin{bmatrix}
            \frac{u_m-c_{x}}{f_{x}},   
            \frac{v_m-c_{y}}{f_{y}}
        \end{bmatrix}^\mathsf{T} = \begin{bmatrix}\frac{X_m}{Z_m},     \frac{Y_m}{Z_m}\end{bmatrix}^\mathsf{T}. 
\end{equation} 
%  
% 
%$\mathbf{K}_{c}$ 
This can be interpreted as a 
point lying on the plane with $Z=1$. 
The generalisability and accuracy of the networks trained with the canonicalised 2D keypoints are evaluated in Sec.\ref{sec:experiments}. 

\subsection{Target Pose Estimation}\label{sec:TPNet} 
After pre-processing the 2D joint keypoints  (Sec.~\ref{ssec:canonicalisation}), the inputs are fed to the target pose estimation network (TPNet) that outputs the global target pose $\mathbf{\hat{q}} \in \mathbb{R}^{n+1}$ and binary labels for the contact states, \textit{i.e.}, toes and heels $\mathbf{b} \in \{0,1\}^{4}$, see Fig. \ref{fig:overview} for the overview. 
TPNet consists of two 1D convolution-based network modules ($\mathcal{C}_{T}$ and $\mathcal{C}_{P}$) that consider temporal information. 
Network $\mathcal{C}_{P}$ first estimates the joint angles and global orientation of the character without the root translation, which is denoted by $\mathbf{\hat{q}}_{rr}$, and foot contact labels $\mathbf{b}$ in the scene; 
$\mathbf{\hat{q}}_{rr}$ is further processed by the forward kinematics layer $f(\cdot)$ to obtain the root-relative 3D joint keypoints $\mathbf{P}_{rr}$ with bone lengths in Cartesian coordinates in the absolute scale.  
Network $\mathcal{C}_{T}$ takes as input $\mathbf{P}_{rr}$ and $\mathbf{K}_{c}$, and outputs the global translation of the character $\mathbf{\hat{q}}_{\text{trans}}$.  
At the end, we obtain global 3D skeleton pose $\mathbf{\hat{q}}$ which is further employed as a target pose of the PD controller (Sec.~\ref{sssec:set_PD}). 
All the networks in TPNet are composed of four residual blocks with 1D convolution layers with window size 10. 
Note that our networks accept only past and current frames with no access to future frames, hence compatible with real-time applications.  

\subsection{Dynamic Cycle}\label{ssec:DynCycle} 
In this section, we elaborate the dynamic cycle of our framework where we control the human character considering dynamics quantities: $\mathbf{M}$, $\mathbf{J}$ and $\mathbf{h}$ are analytically estimated in each simulation step using the current pose $\mathbf{q}_{0}$ and the velocity  $\dot{\mathbf{q}}_{0}$ \cite{featherstone2014rigid}.

\subsubsection{Force Vector Computation by a Neural PD  Controller}\label{sssec:set_PD} 
PD controllers enable motion tracking with a kinematic character while maintaining a smooth motion. 
They are hence widely used in robotics and physics-based animation research \cite{putri2018gait,levine2012physically,sugihara2006gravity,Lee2019,chentanez2018physics}. 
Our framework also utilises a PD controller to compute the internal force vector $\boldsymbol{\tau}$ of the character. 
However, the smoothing properties of PD controller can cause motion delay in the presence of fast motions if the gain values are not optimal. 
The motion delay is especially apparent when the results are shown reprojected to the input views. 
This issue arises from fixing the gains which adjust the PD controller's sensitivity to the pose and velocity error \cite{shimada2020physcap}. 

% 
% \setlength{\columnsep}{8pt}%
% \begin{wrapfigure}{r}{0.25\textwidth}
%   \begin{center}
%     \includegraphics[width=0.23\textwidth]{images/friction_cone.pdf}
%   \end{center}
%   \caption{Schematic visualisation of the friction cone and the ground reaction force at the foot-floor contact position.}\label{fig:friction_cone} 
% \end{wrapfigure}

% 
Similarly to \citet{chentanez2018physics}, we dynamically change the gain coefficients depending on the target and current skeleton poses by our dynamics network (DyNet). 
The latter significantly mitigates the motion delay compared to the existing methods while keeping the motions smooth. 

Our DyNet accepts the target pose $\mathbf{\hat{q}}$, the  current pose $\mathbf{q}_{0}$, the current velocity $\dot{\mathbf{q}}_{0}$, the mass matrix $\mathbf{M}$ and the current pose error $\mathbf{e}_{\text{PD}} = d(\mathbf{\hat{q}}, \mathbf{q}_{0}) \in \mathbb{R}^{n}$, and outputs gain parameters  $\mathbf{k}_{p} \in \mathbb{R}^{n}$ of the PD controller along with the offset forces $\alpha \in \mathbb{R}^{n}$ for each DoF.  
The error function $d(\cdot)$ computes entry-wise differences between $\mathbf{\hat{q}}$ and $\mathbf{q}_{0}$ for the entries that represent the root orientation, we compute the quaternion difference.
Since we provide $\mathbf{\hat{q}}$ and $\mathbf{q}_{0}$, their residual information, \textit{i.e.}, $\mathbf{e}_{\text{PD}}$, is not the essential input for the network. 
However, similar to \cite{bergamin2019drecon}, we observed that explicitly providing the current error to DyNet leads to a much faster loss convergence.
Therefore, we include $\mathbf{e}_{\text{PD}}$ as one of the inputs to DyNet. 
The outputs of TPNet and DyNet are used to compute the force vector $\boldsymbol{\tau}$ following the PD controller rule with the compensation term $\mathbf{h}$\footnote{In literature, this is known as PD controller with force compensation \cite{yang2010pd}.}.: 
\begin{equation} \label{eq:PD_controller}
 \mathbf{\boldsymbol{\tau}}= \mathbf{k}_{p}\circ(\mathbf{\hat{q}}-\mathbf{q_{0}})-\mathbf{k}_d\circ\dot{\mathbf{q}}_{0} + \alpha + \mathbf{h}, 
\end{equation}
where ``$\circ$'' denotes Hadamard matrix product.  $\mathbf{h}$ represents the sum of gravity, centripetal and Coriolis forces, which are analytically computed.

\subsubsection{Ground Reaction Force Estimation}\label{sssec:GRF_estimation}

In real world, external forces are required to control the center of gravity of a human body. 
In other words, for the motion to be physical, the global translation and rotation of the character need to be controlled by external forces such as ground reaction forces obtained from the contact positions. 
On the other hand, the character motion can be controlled to match the pose of the subject in the scene using the force vector $\boldsymbol{\tau}$. 
However, $\boldsymbol{\tau}$ contains direct linear and rotational force applied on the root position as elaborated in Sec.~\ref{sssec:set_PD}. 

We thus train the ground reaction force estimation network (GRFNet) to minimise the (virtual) force applied directly on the root, trying to explain the global motion by the ground reaction force $\boldsymbol{\lambda}$ as much as possible. 
Let $\boldsymbol{\tau}_{\text{root}}\in \mathbb{R}^{6}$ be the force vector corresponding to the root position  (\textit{i.e.}, the first six elements of  $\boldsymbol{\tau}$). 

Then, the main objective function for training GRFNet reads: 
\begin{equation} \label{eq:contact_force}
\mathcal{L}_{\text{force}} = \left\|\boldsymbol{\tau}_{\text{root}}-\mathbf{J}_{1}^{\mathsf{T}} \mathbf{G} \boldsymbol{\lambda}\right\|^{2}_{2},
\end{equation}
where $\mathbf{J}_{1}^{\mathsf{T}}$ denotes the first six rows of $\mathbf{J}^{\mathsf{T}}$ corresponding to the root configuration. 
Minimising \eqref{eq:contact_force} encourages the network to estimate $\boldsymbol{\lambda}$ which explains the forces applied on the root position by GRF. 
\setlength{\columnsep}{8pt}%
\begin{wrapfigure}{r}{0.25\textwidth}
  \begin{center}
    \includegraphics[width=0.23\textwidth]{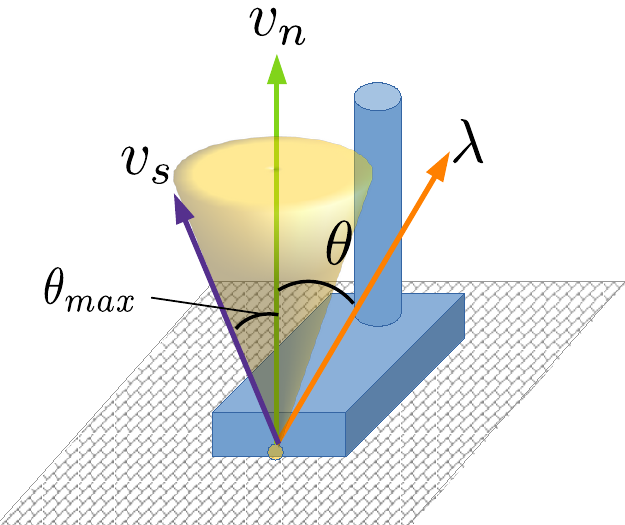}
  \end{center}
  \caption{Schematic visualisation of the friction cone and the ground reaction force at the foot-floor contact position.}\label{fig:friction_cone} 
\end{wrapfigure}

However, the direction of the contact force does not only  depend on \eqref{eq:contact_force}. 
Therefore, we also introduce the friction constraint $\mathcal{F}$ for $\boldsymbol{\lambda}$ to be physically plausible. 
The estimated $\boldsymbol{\lambda}$ needs to be inside of the so-called friction cone which is defined by the friction coefficient $\mu=0.8$ together with the normal and tangential directions of the contact position. 
The friction-cone constraint is defined as follows: 
\begin{equation} \label{eq:friction_constraint}
\mathcal{F}^{\ell}=\left\{\boldsymbol{\lambda}^{\ell} \in \mathbb{R}^{3} \,\Big|\, \lambda_{n}^{\ell}>0, \left \| \lambda_{t}^{\ell}\right \|_{2} \leq \mu \lambda_{n}^{\ell}\right\}, 
\end{equation} 
where $\ell$ represents the identifier of the link where contact force is applied; $\lambda_{n}$ and $\lambda_{t}$ represent the normal and tangential component of $\lambda$, respectively.
We next extend Eq.~\eqref{eq:friction_constraint} by integrating the objective function of GRFNet \eqref{eq:contact_force} in it:  
\begin{equation}\label{eq:friction_constraint_neural} 
  \mathcal{L}_{\text{cone}} = 
    \begin{cases} 
     \left \| \theta \right \|^{2}_{2}, & \mbox{if } \theta > \theta_{\text{max}}, \\ 
     0, & \mbox{else},  
    \end{cases} 
\end{equation} 
where $\theta_{\text{max}}$ is the angle between the normal vector $v_{n}$ of the contact position and a vector $v_{s}$ that lies on the surface of the friction cone, and
$\theta$ is the angle between $v_{n}$ and $\boldsymbol{\lambda}$, see Fig.~\ref{fig:friction_cone}. 
Next, we introduce a temporal smoothness regulariser for the ground reaction force $\boldsymbol{\lambda}$: 
\begin{equation} 
\mathcal{L}_{\text{smooth}} = \left\|\boldsymbol{\lambda}-\boldsymbol{\lambda}_{\text{pre}}\right\|^{2}_{2},
\end{equation}
where $\boldsymbol{\lambda}_{\text{pre}}$ represents the estimated  $\boldsymbol{\lambda}$ in the previous simulation step. 
The final objective function for GRFNet $\mathcal{L}_{\text{GRF}}$ is as follows:
\begin{equation} \label{eq:GRF_loss}
\mathcal{L}_{\text{GRF}} = \mathcal{L}_{\text{force}}+\mathcal{L}_{\text{smooth}}+\mathcal{L}_{\text{cone}}.
\end{equation}

\subsubsection{Forward Dynamics}\label{ssec:FD} 
To introduce the laws of physics in our 3D motion capture algorithm, we embed the forward dynamics layer in our architecture. 
We derive joint accelerations $\ddot{\mathbf{q}}$ from Eq.~\eqref{eq:eom}: 
\begin{equation} \label{eq:fd} 
    \ddot{\mathbf{q}} = \mathbf{M}^{-1}(\mathbf{q}) \big(\boldsymbol{\tau}^{*} +  \mathbf{J}^{\mathsf{T}} \mathbf{G}\boldsymbol{\lambda} - \mathbf{h}\big), 
\end{equation} 
where $\boldsymbol{\tau}^{*} = \boldsymbol{\tau} - \mathbf{J}^{\mathsf{T}}  \mathbf{G}\boldsymbol{\lambda}$. 
In this formulation, $\boldsymbol{\tau}^{*}$ expresses the minimised direct root actuation with contact force compensation for each joint torque. 
This forward dynamics layer returns $\ddot{\mathbf{q}}$  considering mass matrix of the body $\mathbf{M}$, internal and external forces, gravity, Coriolis and centripetal forces encompassed in $\mathbf{h}$. 

\subsubsection{Constrained Pose Update}\label{sssec:PUHC} 
In this step, we update the character's pose using the estimated accelerations $\ddot{\mathbf{q}}$ through the differentiable optimisation layer to prevent foot-floor penetration. 
Given $\ddot{\mathbf{q}}$ in the skeleton frame and the  simulation time step $\Delta t$, the velocity  $\dot{\mathbf{q}}$ and the kinematic 3D pose $\mathbf{q}$ are updated using the finite differences: 
\begin{equation} \label{eq:findif} 
\begin{aligned} 
    &\dot{\mathbf{q}}^{i+1}=\dot{\mathbf{q}}^{i}+\Delta t\,\ddot{\mathbf{q}}^{i},\\ 
    &\mathbf{q}^{i+1}= \mathbf{q}^{i} +\Delta t\,\dot{\mathbf{q}}^{i+1}, 
\end{aligned} 
\end{equation}
where $i$ denotes the simulation step identifier. 
To prevent foot-floor penetration, we introduce the differentiable optimisation layer following the formulation of \cite{agrawal2019differentiable}. 
This custom neural network layer solves a specific optimisation problem for each forward pass and returns its derivatives for each backward pass. 
More specifically, we update the velocity in the skeleton frame $\dot{\mathbf{q}}$ solving the optimisation below:
\begin{equation}\label{eq:difoptim} 
\begin{matrix}
    \underset{\mathbf{\dot{\mathbf{q}}^{*}}}{\min } \left\| \dot{\mathbf{q}}^{*}-\dot{\mathbf{q }} \right\|, \;  
    \text{s.t.}\;\mathbf{r}_{n}^{c} > 0, 
    \end{matrix}
\end{equation} 
where $\mathbf{r}_{n}^{c}$ represents the linear velocity of the contact position along the normal direction of the floor. Velocity vector $\mathbf{r}^{c}$ is computed as follows: 
\begin{equation} \label{eq:lin_vel} 
\mathbf{r}^{c} = \mathbf{T}(\mathbf{J}\dot{\mathbf{q}}), 
\end{equation} 
where $\mathbf{T}(\cdot)$ is the transformation from the camera frame to the floor frame of reference. 
After solving \eqref{eq:difoptim}, the estimated  $\dot{\mathbf{q}}^{*}$ is substituted as  $\dot{\mathbf{q}}$ in Eq.~\eqref{eq:findif}. 
The dynamic cycle introduced in this section is iterated $k = 6$ times. 
After the iterations are complete, we obtain the final physically-plausible 3D character's pose $\mathbf{q}$.  

\subsection{Network Training}\label{ssec:net_train} 
We pre-train TPNet for a more stable training of the whole architecture. 
Such pre-training is advantageous due to two reasons. 
First, estimating joint angles from 2D joint keypoints leads to ambiguities in bone orientations \cite{shi2020motionet}.  
Second, controlling the dynamic character in 3D by estimated forces to match the subject's pose only from 2D joint keypoints is an ill-posed problem. 
The network $\mathcal{C}_{P}$ in TPNet is pre-trained with the following objective loss function: 
\begin{equation}\label{eq:TP_objective} 
\begin{aligned} 
& \mathcal{L}_{\mathcal{C}_{P}} = \mathcal{L}_{\text{3D}}(\mathbf{\hat{q}}) +  \mathcal{L}_{\text{2D}}(\mathbf{\hat{q}}) + \mathcal{L}_{\text{ori}}(\mathbf{\hat{q}}_{\text{ori}})+\mathcal{L}_{\text{irr.}}(\mathbf{\hat{q}}) + \mathcal{L}_{b}(\mathbf{b}).  
\end{aligned} 
\end{equation} 
The main 3D loss $\mathcal{L}_{\text{3D}}$ is defined as follows: 
\begin{equation} 
 \mathcal{L}_{\text{3D}}(\mathbf{\hat{q}}) = \left\|  f(\mathbf{\hat{q}}) - \mathbf{p}'_{\text{3D}} \right\|^{2}_{2},
\end{equation}
where $f(\cdot)$ and $\mathbf{p}'_{\text{3D}}$ are forward kinematics function and ground-truth 3D joint keypoints,  respectively. 
The loss $\mathcal{L}_{\text{2D}}$ stands for the 2D reprojection error % 
\begin{equation} 
 \mathcal{L}_{\text{2D}}(\mathbf{\hat{q}}) = \left\| \Pi(f(\mathbf{\hat{q}})) - \mathbf{p}'_{\text{2D}} \right\|^{2}_{2},
\end{equation}
where $\Pi(\cdot)$ and $\mathbf{p}'_{\text{2D}}$ are the perspective projection operator and ground-truth 2D joint keypoints normalised by the image size, respectively. 
The loss $\mathcal{L}_{\text{ori}}$ is added for the  supervision of the global root orientation represented by a  quaternion: 
\begin{equation} 
 \mathcal{L}_{\text{ori}}({\mathbf{\hat{q}}}_{\text{ori}}) = \left\| \mathbf{\hat{q}}_{\text{ori}}  \ominus  \mathbf{q}'_{\text{ori}}\right\|^{2}_{2},
\end{equation}
where $\mathbf{q}'_{\text{ori}}$ is the ground-truth root orientation in quaternion parame-trisation, and ``$\ominus$'' denotes a difference computation after converting the quaternion into a rotation matrix. 
The loss $\mathcal{L}_{\text{irr.}}$ keeps the estimated joint angles in a reasonable range: 
\begin{equation} 
  \mathcal{L}_{\text{irr.}}({\mathbf{\hat{q}}}) =  \sum^{40}_{i=1}\Psi(\mathbf{\hat{q}}_{i}), \,\text{with}\,  
\end{equation} 
\begin{equation} 
  \Psi(\mathbf{\hat{q}}_{i}) = 
    \begin{cases} 
     (\mathbf{\hat{q}}_{i}-\psi_{\text{max},i})^{2}, & \mbox{if } \mathbf{\hat{q}}_{i} > \psi_{\text{max},i} \\ 
     (\psi_{\text{min},i}-\mathbf{\hat{q}}_{i})^{2}\;, & \mbox{if } \mathbf{\hat{q}}_{i} < \psi_{\text{min},i} , \\
     0\;\;\;\;\;\;\;\;\;\;\;\;\;\;\;\;\;\;\;, & \mbox{otherwise},
    \end{cases} 
\end{equation}
where $\mathbf{\hat{q}}_{i}$ denotes the joint angle of the $i$-th joint and [$\psi_{\text{min},i}$, $\psi_{\text{max},i}$] defines the reasonable angle range for the $i$-th joint. 
Term $\mathcal{L}_{b}$ is the binary cross entropy loss to train the network for estimating correct foot contact states in the scene: 
\begin{equation} 
 \mathcal{L}_{b}(\mathbf{b}) = -\sum_{i=1}^{4}b'_{i}\log(b_{i})+(1-b'_{i}) \log(1-b_{i}),
\end{equation}
where $b'_{i}$ and $b_{i}$ are the ground-truth contact label and predicted contact probability on $i$-th joint, respectively. 
The $\mathcal{C}_{T}$ module of TPNet is trained with the 3D  translation loss: 
\begin{equation} 
 \mathcal{L}_{C_{T}}(\mathbf{\hat{q}}_{\text{trans}}) = \left\| \mathbf{\hat{q}}_{\text{trans}} - \mathbf{q}'_{\text{trans}} \right\|^{2}_{2},
\end{equation} 
where $\mathbf{q}'_{\text{trans}}$ denotes the ground-truth translation in 3D space.
After pre-training $\mathcal{C}_{P}$ and $\mathcal{C}_{T}$ with  $\mathcal{L}_{C_{T}}$ and  $\mathcal{L}_{C_{P}}$, we train DyNet with the following loss: 
\begin{equation} 
 \mathcal{L}_{\text{Dyn}}(\mathbf{q}) = \left\| \mathbf{q} - \mathbf{\hat{q}} \right\|^{2}_{2}+ \varphi \left\| \tau \right\|^{2}_{2},
\end{equation} 
where $\mathbf{q}$ is the final, physically-plausible 3D pose passed through the differentiable physics model and $\varphi = 10^{-6}$ is the weight of the regularisation term of $\boldsymbol{\tau}$. 
The first term of $\mathcal{L}_{\text{Dyn}}$ enforces the character to catch up with the target pose with the mitigated motion delay by dynamically estimating the gain parameters of the PD controller. 
The second term of $\mathcal{L}_{\text{Dyn}}$ prevents   overshooting of the PD controller output. 
The GRFNet is trained with Eq.~\eqref{eq:GRF_loss}  (Sec.~\ref{sssec:GRF_estimation}). 
After pre-training all the networks until convergence, all the networks are trained jointly with the corresponding objective functions with an early stopping strategy. We use Adam optimiser with a learning rate $3.0\times 10^{-6}$ for the pre-training, and $3.0\times 10^{-7}$ for the joint training.

\subsection{Adaptations for In-the-Wild  Recordings}\label{ssec:in_the_wild} 
Our framework allows finetuning the networks with 2D  annotations only using the 2D reprojection loss. 
Such adjustment of the network weights is especially effective for in-the-wild recordings which differ from the training samples in many aspects (\textit{e.g.}, in the background, lighting conditions or camera poses). 
We use the estimated 2D joint keypoints from OpenPose \cite{openpose1,openpose2,openpose3,openpose4} as pseudo ground-truth 2D annotation to train our network, see Fig.~\ref{fig:finetune} for the results of the ablative study and our supplementary video for visual comparisons of the results with and without finetuning.  

\subsection{Applications}\label{ssec:applications} 
Since our framework estimates the global translation, root orientation and joint angles, virtual characters can be directly animated using the output of our method.  
We can also visualise the estimated torques and ground reaction forces that explain the motion in the scene, see Fig.~\ref{fig:teaser}-(right) for an example. 
The purple vectors represent the estimated ground reaction  forces, and more saturated green hue on the links represents  stronger torques applied on the child joints. 
% 
%
\begin{comment} 
\begin{equation}
    \mathbf{K}_{c} = 
    \bigg(
        \begin{bmatrix} 
        \frac{u_1-c_{x}}{f_{x}}, \frac{v_1-c_{y}}{f_{y}} 
        \end{bmatrix} \\
        \begin{bmatrix} 
        \frac{u_2-c_{x}}{f_{x}},\frac{v_2-c_{y}}{f_{y}} 
        \end{bmatrix}
        \hdots 
        \begin{bmatrix} 
        \frac{u_M-c_{x}}{f_{x}},\frac{v_M-c_{y}}{f_{y}} 
        \end{bmatrix} 
        \bigg)^\mathsf{T} 
\end{equation} 
\begin{equation}\label{eq:canonical_keys_testing}
\mathbf{K}^{s}_{c} =  \begin{bmatrix}
\frac{u-c_{x}}{f_{x}}\\ 
\frac{v-c_{y}}{f_{y}}\\  
\end{bmatrix},
\end{equation}
\end{comment}

\begin{table*}[ht!]
\setlength\heavyrulewidth{0.20ex} 
\aboverulesep=0ex
\belowrulesep=0ex
\center
\caption{\label{tab:error_3D} Comparisons of 3D joint position errors on DeepCap \cite{deepcap}, Human 3.6M  \cite{ionescu2013human3} and  MPI-INF-3DHP\cite{mono-3dhp2017} datasets. 
From the kinematic-based algorithm class, we compare  with VNect \cite{VNect_SIGGRAPH2017}, HMR  \cite{hmrKanazawa17}, HMMR  \cite{humanMotionKanazawa19}, VIBE  \cite{kocabas2020vibe} and MotioNet  \cite{shi2020motionet}. 
From the physics-based algorithm class, we compare our method with PhysCap \cite{shimada2020physcap}. 
``$\dagger$'' denotes physics-based algorithms, otherwise a kinematic algorithm.  
``$*$'' denotes MotioNet with causal convolutions which does not have access to the future frames, \textit{i.e.}, the similar problem set as our approach. 
Our approach shows competitive results with kinematic approaches, and outperforms physics-based approaches with a big margin in most  scenarios. For DeepCap dataset, the numbers on left and right of our approach represent the 3D accuracy with and without training on DeepCap dataset, respectively.
} 
 \scalebox{0.90}{
 \begin{tabular}{ c l c c c c c c c c c c c }\toprule
 
 				 &    &  \multicolumn{3}{c}{DeepCap}     &&  \multicolumn{3}{c }{Human 3.6M}  &&  \multicolumn{3}{c }{MPI-INF-3DHP}  \\  
 				 \cmidrule(lr){3-5} \cmidrule(lr){7-9} \cmidrule(lr){11-13} 
				  &   &  MPJPE [mm]$\downarrow$ &  PCK [\%]$\uparrow$  &  AUC [\%]$\uparrow$  &&  
				 MPJPE [mm]$\downarrow$ &  PCK [\%]$\uparrow$ &  AUC [\%]$\uparrow$ &&  MPJPE [mm]$\downarrow$ &  PCK [\%]$\uparrow$  &  AUC [\%]$\uparrow$ \\  \addlinespace[2pt] \midrule
 
 \multirow{6}{*}{\rotatebox[origin=c]{90}{Procrustes}}    
        &  Ours$\dagger$		    &   \textbf{52.6}/63.6 & \textbf{97.3}/95.9  &  \textbf{67.1}/60.1  &&  58.2  & 96.1 & 64.4 && 99.1 & 85.5 & 42.7  \\
         &  PhysCap$\dagger$			    & 68.9     &  95.0  & 57.9    && 65.1   & 94.8    & 60.6 && 104.4 & 83.9 & 43.1 \\ 
       &  MotioNet*    & 123.0   & 73.0  &  31.0  && 59.1  &  -  & -  &&  -  & -   &  -   \\  
      &  VIBE		    & 80.1   & 93.3  & 50.1   && \textbf{41.5}  & - & - && \textbf{63.4} & - & -  \\  
      
     &  VNect   &  68.4 & 94.9	 &    58.3  && 62.7   & 95.7     & 61.9  && 104.5 & 84.1 & 43.2  \\
     &   HMR    & 77.1    & 93.8  & 52.4     && 54.3  &  \textbf{96.9}     & \bf{66.6}  && 87.8& \bf{87.1} & \bf{50.9}  \\ 
     &  HMMR    & 75.5		& 93.8	  & 53.1     && 55.0   & 96.6 & 66.2  && 106.9  &79.5 & 44.8  \\ \hline
  \multirow{6}{*}{\rotatebox[origin=c]{90}{no Procrustes}}    
          &  Ours$\dagger$		    &  \textbf{72.7}/88.6  &  \textbf{92.6}/85.7 &  \textbf{55.3}/47.4  &&  76.5 &  \textbf{89.5}   & \textbf{55.0} && 134.5  & 69.8  & 30.2  \\
           &   PhysCap$\dagger$		    & 113.0     & 75.4   & 39.3     && 97.4   & 82.3     & 46.4 && 122.9&72.1 &35.0  \\ 
       &  MotioNet*	    & 257.4   & 33.0  &  13.3  &&  -  &  - & - &&  -  &  -  &  -   \\  
      &  VIBE		    &  96.7  & 85.9  & 42.4   &&  \textbf{65.9} & - & - && \textbf{97.7}  &  - &  -  \\   
    
    &   VNect    & 102.4		& 80.2	 &  42.4 && 89.6   & 85.1     & 49.0 &&  120.2 & \bf{74.0} & \bf{36.1} \\
    &   HMR	    & 113.4     & 75.1  & 39.0   &&  78.9    & 88.2     &  54.1 && 130.5 & 69.7 & 35.7 \\
     &  HMMR     &  101.4 	&  81.0   & 42.0    && 79.4   & 88.4     & 53.8  && 174.8 & 60.4 & 30.8 \\ \bottomrule 
\end{tabular}
}
\end{table*}
  
\section{Experiments}\label{sec:experiments}

We evaluate our physionical approach for monocular 3D  human motion capture on Human 3.6M  \cite{ionescu2013human3}\footnote{All experiments and training using Human 3.6M were conducted at MPII.}, MPI-INF-3DHP \cite{mono-3dhp2017}, DeepCap \cite{deepcap} as well as newly recorded sequences. We first provide implementation details (Sec.~\ref{ssec:implementation}) and then show qualitative results (Sec.~\ref{ssec:quantitative_results}) as well as the quantitative outcomes (Sec.~\ref{ssec:quantitative_results}). 

\subsection{Implementation}\label{ssec:implementation} 
Our neural networks are implemented using PyTorch \cite{NEURIPS2019_9015} and Python 3.7. Adam optimiser was used to train them.
For the computation of dynamics quantities, we use \textit{Rigid Body Dynamics Library} \cite{Felis2016}. 
For the implementation of the differentiable optimisation layer we use \cite{cvxpylayers2019}, and  
\textit{Pybullet} \cite{coumans2016pybullet} for visualisation purposes. 
Our approach is evaluated on a workstation with 
32 GB RAM, AMD EPYC 7502P 32-Core Processor and NVIDIA QUADRO RTX 8000. 
\subsection{Network Details}\label{ssec:networkdetails} 
Our implementation of $\mathcal{C}_{T}$ and $\mathcal{C}_{P}$ is composed of 4 1D-convolution-based residual blocks which consider temporal information. GRFNet consists of $4$ fully-connected layers with ReLU activation functions excepting the output layer. DyNet forms a two-headed network to estimate the gain parameters of PD controller and offset forces. Two fully-connected layers and ReLU activation functions are used for its hidden layers. One fully-connected layer is used for its output layer followed by Sigmoid and Tanh activation functions for each head of the network. See Appendix A for the network details.

\subsection{Quantitative Results}\label{ssec:quantitative_results} 

\begin{table}
\center
\caption{\label{tab:global_translation} Global 3D translation error on DeepCap dataset \cite{deepcap}. Note that our networks are trained on Human3.6M \cite{ionescu2013human3} and MPI-INF-3DHP \cite{mono-3dhp2017}, and \textit{not} trained on DeepCap dataset \cite{deepcap}.
}  
\scalebox{0.83}{
\begin{tabular}{c c c c c c c}\bottomrule 
				  & Ours &  \makecell{Ours w/o \\  $C_{T}$  module} &  \makecell{Ours w/o\\ input cano.} &  PhysCap  & VNect   &  VIBE\\ \midrule  
       MPJPE [mm]$\downarrow$ & \bf{62.6}    &  68.7 & 105.0& 110.5 & 112.6  & 244.5  \\  \bottomrule
\end{tabular}
}
\end{table} 

In this section, we compare our method with other related kinematic-based methods, \textit{i.e.}, VNect  \cite{VNect_SIGGRAPH2017}, HMR  \cite{hmrKanazawa17}, HMMR  \cite{humanMotionKanazawa19}, VIBE  \cite{kocabas2019vibe} and MotioNet  \cite{shi2020motionet}, as well as the recent physics-based method PhysCap \cite{shimada2020physcap} on benchmark  datasets \cite{ionescu2013human3, mono-3dhp2017, deepcap}.  

We follow the evaluation methodology proposed in  \cite{shimada2020physcap} which suggests comparisons of monocular 3D human motion capture using an extended set of metrics.  
Along with the standard root-relative 3D joint position accuracy metrics, \textit{i.e.}, mean per-joint position error (MPJPE) [mm] (the lower the better), percentage  of correct keypoints [$\%$] and area under ROC curve (AUC) [$\%$] (the higher the better), 
we report the global 3D translation error and 2D re-projection errors by projecting the estimated 3D joints onto the input and evaluation (unseen) views.  
Reprojection to evaluation views reveals various effects (related to physical implausibility) which are difficult to access based only on root-relative 3D errors and reprojections to the input views.  
Further complementary metrics  measuring the degree of plausibility of the reconstructed poses are Mean Penetration Error (MPE), Percentage of Non-Penetration (PNP) and temporal consistency error. 
MPE evaluates the average distance between the foot and floor when there is actually a foot-floor contact in the scene (lower is better). 
PNP shows the ratio of no foot penetration into the floor (higher reflects higher degree of physical plausibility).  

\paragraph{3D Joint Positions}  
Table \ref{tab:error_3D} summarises the root-relative 3D joint position errors. 
The first and second row blocks report the calculations with and without Procrustes alignment, respectively.  
On Human 3.6M and MPI-INF-3DHP with Procrustes alignment, the accuracy of our method is average among the compared methods. 
On Human 3.6M, we obtain a slightly lower MPJPE than VNect, MotioNet and PhysCap while HMR, HMMR and VIBE achieve the lowest errors in overall. 
On MPI-INF-3DHP, the overall tendency is preserved, though in addition we outperform HMMR. 
On the DeepCap dataset, we report the results of two different variants, \textit{i.e.}, when the networks are trained on DeepCap dataset + Human3.6M + MPI-INF-3DHP (on the left) and Human3.6M + MPI-INF-3DHP without DeepCap dataset (on the right). Even without using DeepCap dataset for training, ours outperforms other tested algorithms. 
Compared with Human 3.6M and MPI-INF-3DHP, DeepCap dataset contains challenging motions such as dance, walking backwards, jumping and running sequences. 
Purely kinematic algorithms tend to fail on these  challenging motions. 
In our case, the magnitudes of inaccuracies are regularised within a reasonable range thanks to the explicit physics model, which results in a lower MPJPE. 

Most of the competing methods overfit to a single  dataset and cannot generalise well to other datasets. Without Procrustes alignment, our approach outperforms all other evaluated methods on DeepCap dataset, and ranks second on Human 3.6M. We consistently outperform the most related methods on DeepCap, Human 3.6M and MPI-INF-3DHP (with Procrustes), which estimate global 3D human poses and can be directly used for virtual character animation. 
This list also includes the physics-based PhysCap, \textit{i.e.}, the most closely related method to ours.  
The high accuracy of purely kinematics methods (in Table  \ref{tab:error_3D}, those are all methods without ``$\dagger$'') on Human 3.6M and MPI-INF-3DHP comes at the price of frequent and sudden  changes in the 3D joint positions, which result in jitters and other artefacts. 
See our supplementary video for the qualitative examples. 

Note that the obvious artefacts such as jitter and foot-floor penetration are not revealed by these conventional metrics, which suggests that considering those alone is not enough to judge the motion quality: they do not draw the complete picture, especially when having computer graphics applications in mind; hence, we report several additional metrics to provide a more comprehensive assessment of the motions. 

\paragraph{Global Translation Errors}  
We also qualitatively compare the accuracy of the global character's root position (translation) on the DeepCap dataset in Table~\ref{tab:global_translation}. Note that we train our method only on Human 3.6M and MPI-INF-3DHP datasets in this experiment, which also evaluates the generalisability of the translation estimator $C_{T}$ trained with the canonical 2D keypoints. 
We also show our ablated models 1) without the $\mathcal{C}_{T}$ module and 2) without the input canonicalisation, in the third and fourth columns, respectively. 
In the third case, instead of using $\mathcal{C}_{T}$, we estimate the global translation by solving a 2D reprojection-based optimisation with gradient descent, given the estimated root-relative 3D pose and 2D joint keypoints.
Without the input keypoint canonicalisation, the performance of our algorithm is significantly decreased compared to our full model. This is because the network overfits to the camera parameters which are observed in the training datasets without the input canonicalisation.
For VIBE---since it does not return a global 3D translation---we apply re-scaling of bone lengths to match the ground-truth bone lengths and likewise solve a reprojection-based optimisation to estimate the global translation which we report in the seventh column. 
We see that even without $C_{T}$ module activated, our method outperforms PhysCap, VNect and VIBE by $75\%$ (for PhysCap) and more (VNect and VIBE). 
See our supplementary video for the qualitative  comparisons.

\begin{table} 
\setlength\heavyrulewidth{0.20ex} 
\aboverulesep=0ex
\belowrulesep=0ex
\center
\caption{\label{tab:jitter} Comparison of temporal smoothness on the DeepCap \cite{deepcap} and Human 3.6M datasets \cite{ionescu2013human3}.  
}
 \scalebox{0.82}{
 \begin{tabular}{ l c c c c c c c }\toprule
	 		  &   & Ours & PhysCap & VNect & HMR & HMMR & VIBE \\   \midrule
   \multirow{2}{*}{DeepCap}	  &  $e_{\text{smooth}}$&\bf{5.8}  & 6.3 & 11.6	    & 11.7     & 8.1  & 7.2  \\ 
      & $\sigma_{\text{smooth}}$      & 8.1 & 4.1     & 8.6	    & 9.0      & 5.1  & 10.1 \\  \midrule
   \multirow{2}{*}{Human 3.6M}  &  $e_{\text{smooth}}$	& \bf{4.5} &  7.2      & 11.2	    & 11.2     & 6.8 & -  \\ 
      & $\sigma_{\text{smooth}}$     &  6.9  &  6.9      & 10.1	    & 12.7     & 5.9 & -  \\  \bottomrule
\end{tabular}
}
\end{table}
  
\begin{table}
\setlength\heavyrulewidth{0.20ex} 
\aboverulesep=0ex
\belowrulesep=0ex
\center
\caption{\label{tab:projection} 2D projection error of a frontal view (input) and side view (non-input) on DeepCap dataset \cite{deepcap}.  
} 
\scalebox{1.0}{
\begin{tabular}{ l c c c c c}\bottomrule
                    &  \multicolumn{2}{c}{Front View}  &   &  \multicolumn{2}{c }{Side View}\\ 
                    \cmidrule(lr){2-3} \cmidrule(lr){5-6}
				    &  $e_{\text{2D}}^{\text{input}}$ [pix]  &  $\sigma_{\text{2D}}^{\text{input}}$  & & $e_{\text{2D}}^{\text{side}}$ [pix]  &  $\sigma_{\text{2D}}^{\text{side}}$  \\ \addlinespace[2pt]  \midrule
      Ours		    &   ~~~\textbf{7.6}  &7.5 & & \textbf{11.5}  & \textbf{13.1}  \\ 
      PhysCap		    & 21.1     & 6.7  & & 35.5     & 16.8   \\ 
      VNect       &  14.3 	& 2.7 & & 37.2     & 18.1 	 \\  \bottomrule    
\end{tabular}
}
\end{table} 

\paragraph{Physical Plausibility Measurement} 
We further evaluate our approach using quantitative measures for the plausibility of the 3D motion. 
Table \ref{tab:jitter} shows the temporal smoothness error $e_{\text{smooth}}$ which is computed as follows  \cite{shimada2020physcap}: 
\begin{equation} \label{eq:jitter} 
\begin{split}
& e_{\text{smooth}} = \frac{1}{Tk}\sum_{t=1}^{T}\sum_{s=1}^{k}\left\|\text{Jit}_{\text{GT}}-\text{Jit}_{X}\right|,\\
& \text{with}~
\text{Jit}_{X} =  \left\| \mathbf{p}^{s,t}_{X} - \mathbf{p}^{s,t-1}_{X}\right\| \;\text{and}\; 
\text{Jit}_{\text{GT}} =  \left\| \mathbf{p}^{s,t}_{\text{GT}} - \mathbf{p}^{s,t-1}_{\text{GT}}\right\|, 
\end{split}
\end{equation} 
where $\mathbf{p}^{s,t}$ represents the 3D position of joint $s$ in the frame $t$; 
$T$ and $k$ denote the total numbers of frames in the input sequence and target 3D joints, respectively. 
Smaller $e_{\text{smooth}}$ means less jitter in the reconstructed 3D motions. 
Our approach shows the lowest $e_{\text{smooth}}$ among all tested methods, followed by physics-based method PhysCap and VIBE and HMMR with temporal constraints (\textit{i.e.},  these methods take several frames as inputs). 
This confirms the significance of our explicit physics model for more physically-plausible results. 

We also report in Table \ref{tab:projection} the 2D reprojection error onto the input views ($e_{\text{2D}}^{\text{input}}$) and side views ($e_{\text{2D}}^{\text{side}}$) that are not used as inputs to the algorithms: 
$\sigma_{\text{2D}}^{\text{input}}$ and $\sigma_{\text{2D}}^{\text{side}}$ represent the standard deviation of $e_{\text{2D}}^{\text{input}}$ and $e_{\text{2D}}^{\text{side}}$, respectively. 
Reprojection onto non-input-view is an expressive operation, since it reveals the artefacts which are not observable from the input view (\textit{e.g.}, body leaning and wrong translation estimation along the depth direction). 
Again, our results lead to the lowest metric among all methods which  suggests that our global 3D motion capture is more physically plausible compared to other methods. 

\begin{table}[t]
\setlength\heavyrulewidth{0.20ex} 
\aboverulesep=0ex
\belowrulesep=0ex
\center
\caption{\label{tab:penetration} Comparison of Mean Penetration Error (MPE) and Percentage of Non-Penetration (PNP) on DeepCap dataset \cite{deepcap}. 
}
\scalebox{1.0}{
\begin{tabular}{ l c c  }\bottomrule
				    & MPE [mm]$\downarrow$   & PNP [\%]$\uparrow$ \\   \midrule
	  Ours		    &  28.9   & 92.3 \\ 
	  Ours w/o HC	    &  29.7   & 89.6 \\ 
      PhysCap		    & \textbf{28.0}       &  \textbf{92.9} \\ 
      VNect       & 39.3	   	& 45.6	 \\   \bottomrule 
\end{tabular}
}
\end{table}

\begin{figure}[t] 
\centering 
\includegraphics[width=\linewidth]{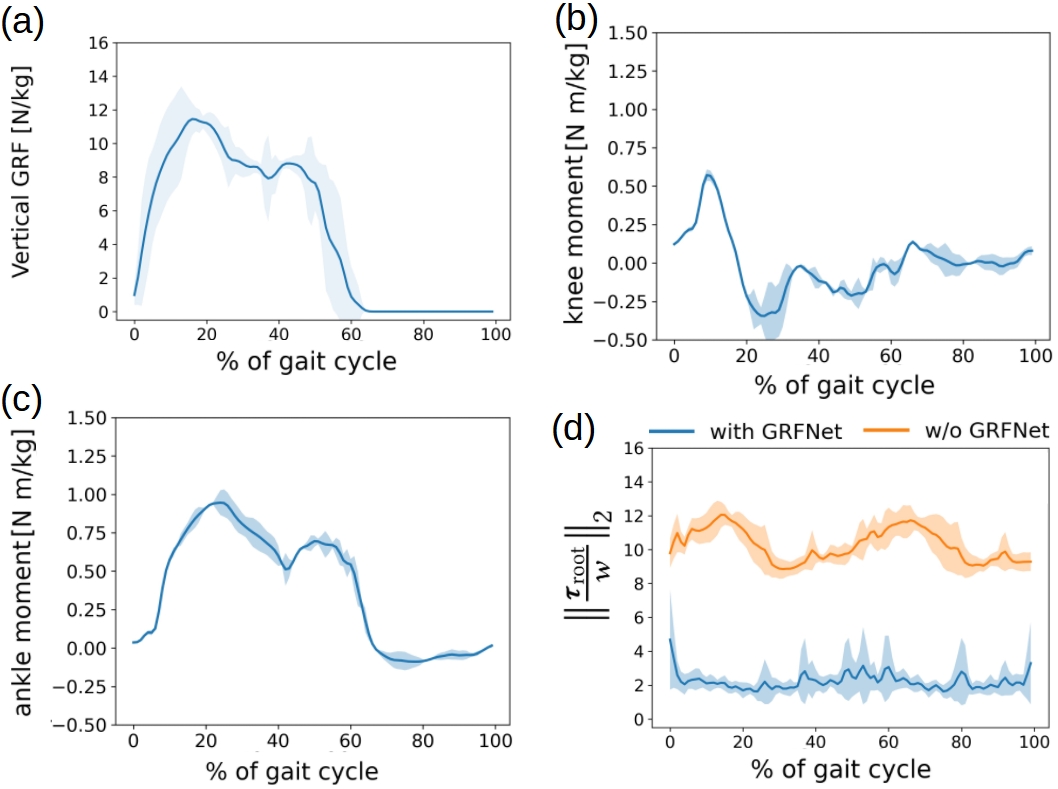} 
\caption{Estimated forces of the walking sequences from the DeepCap dataset. The thick line and coloured area represent the means and standard deviations, respectively. The force graph lies in the reasonable range for walking motion (\textit{cf.}~\cite{Zell2020,shahabpoor2017measurement}), and mostly shows a smooth curve. } 
\label{fig:gait_force} 
\end{figure}

Finally, Table \ref{tab:penetration} reports the physical plausibility measurement for foot-floor penetration. Our result is on par with PhysCap which introduces hard constraint to prevent foot-floor penetration, followed by the purely kinematic method VNect. We also show our ablated model without the hard constraint layer (Sec.~\ref{sssec:PUHC}). Compared to it, our full architecture shows better performance in terms of the foot-floor penetration metric.

\begin{figure*}[t] 
\centering 
\includegraphics[width=\linewidth]{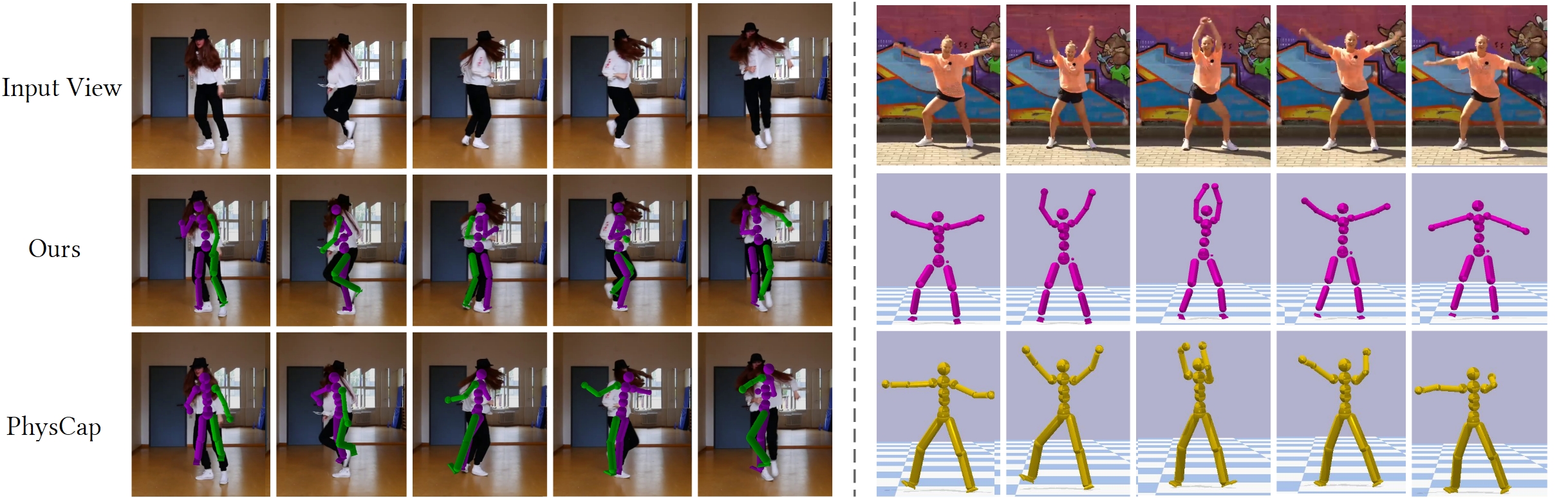} 
\caption{Qualitative comparisons of methods with physics-based constraints on videos with fast motions.  
While having a consistently improved accuracy on general motions compared to PhysCap, our approach can capture significantly faster motions as it learns motion priors and the associated gains of the neural PD controller from data. 
} 
\label{fig:physcap_comparison} 
\end{figure*}

\paragraph{GRF Function} 

In Fig.~\ref{fig:gait_force}, we plot the forces estimated by our physionical algorithm for the walking motion from the DeepCap dataset. 
The thick lines and coloured areas represent the mean values and standard deviations, respectively. 
In Figs.~\ref{fig:gait_force}-(a), (b) and (c), we show the estimated GRF along the vertical direction and joint torques of knee and ankle, respectively. 
The curve is smooth and is in a reasonable range for walking motions. 
Interested readers are referred to \cite{shahabpoor2017measurement,Zell2020} for a visual comparison with ground-truth GRF curves for an exemplary walking sequence obtained with force plates. 
Note that our approach accepts only a single 2D image sequence as input  and does not require any ground-truth forces for its training unlike \cite{Zell2020}. 
In Fig.~\ref{fig:gait_force}-(d), we show an ablative study of GRFNet. 
As elaborated in Sec.~\ref{sssec:GRF_estimation}, GRFNet minimises the presence of unnatural virtual forces directly applied on the character's root joint $\boldsymbol{\tau}_{\text{root}}$ and tries to explain the root motion by the GRF only, as much as possible. 
We report $\left \| \frac{\boldsymbol{\tau}_{\text{root}}}{w}\right  \|_{2}$ for walking cycles, where $w$ is the character's weight. 
Without GRFNet, the magnitude of the virtual force acting directly on the root is 
${\sim}5$ times higher compared to the case with the former. 
This suggests that GRFNet helps to estimate more physically-plausible forces in the proposed framework.

\begin{figure}[t] 
\centering 
\includegraphics[width=\linewidth]{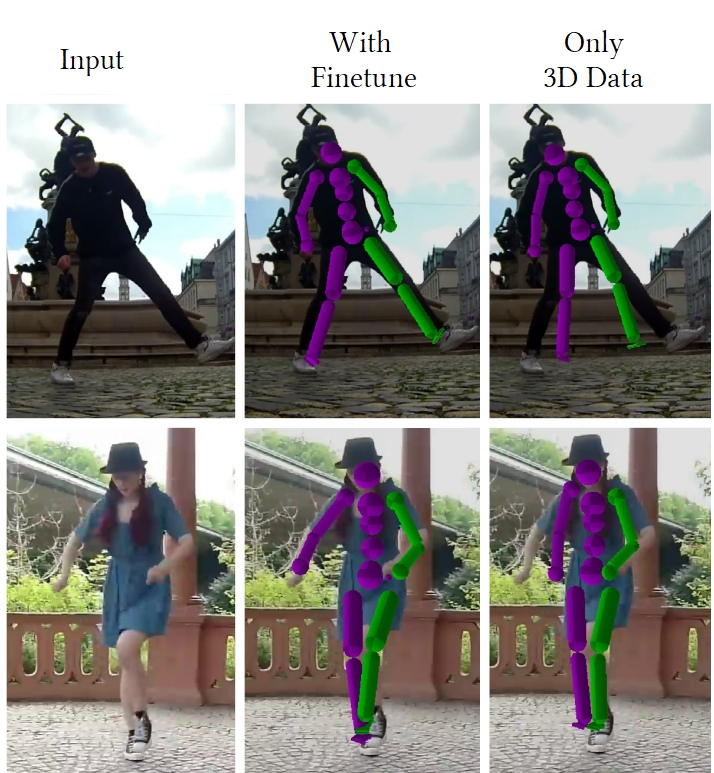} 
\caption{\label{fig:finetune}The accuracy of our method with 
finetuning using additional 2D annotations improves for in-the-wild sequences, compared to training using 3D data only.  
}  
\end{figure} 
 
\begin{figure*}[ ] 
\centering 
\includegraphics[width=20.55cm,angle=-90,origin=c]{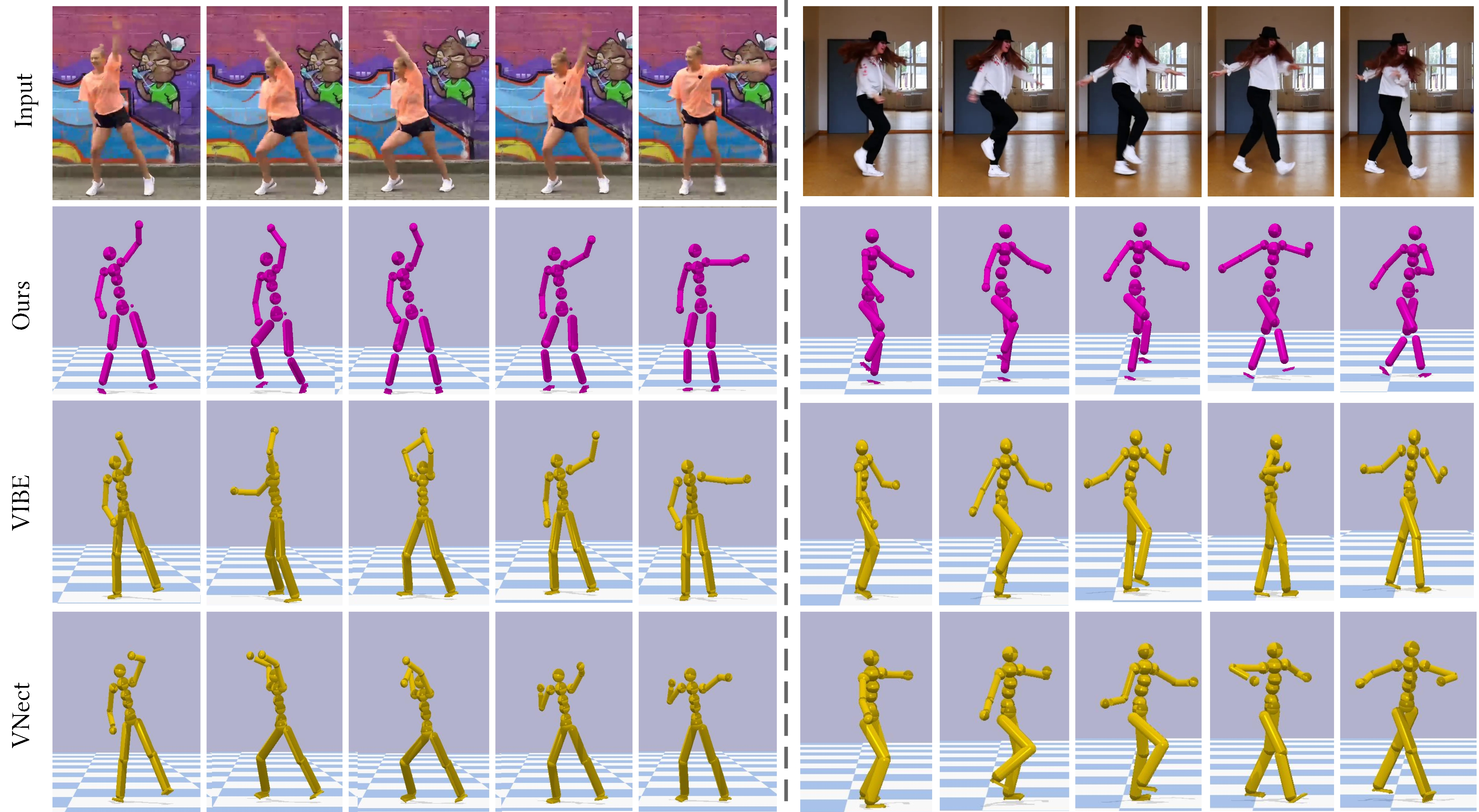} 
\caption{Results of our method compared to  purely-kinematic methods VIBE (3D human pose  estimation) \cite{kocabas2020vibe} and VNect (3D human motion capture) \cite{VNect_SIGGRAPH2017}. 
Our reconstructions are more temporally smooth, whereas the competing methods show frame-to-frame jitter along all axes. 
See our supplementary video for dynamic visualisations. 
} 
\label{fig:kinematic_comparison} 
\end{figure*}
\subsection{Qualitative  Results}\label{ssec:qualitative_results} 
We further show results on multiple in-the-wild sequences. 
All in all, we observe that our physionical method outputs temporally-consistent global 3D human poses which not only accurately project to the input views but which also look  physically plausible when observed from arbitrary views in the 3D space. 
Our reconstructed 3D motions show significantly mitigated physically-implausible artifacts such as spurious global translational variations along the depth dimension, foot-floor penetration and jitters, see our supplemental video for the qualitative results. 
We qualitatively compare our method with the most related work PhysCap \cite{shimada2020physcap} in  Fig.~\ref{fig:physcap_comparison}. 
It is noticeable that our method catches up with fast motions with significantly mitigated motion delay thanks to the 
learned PD controller gain values for different motion types (see Fig.~\ref{fig:physcap_comparison}-(left)). 
PhysCap struggles to reconstruct correct 3D motions when fast motion appears due to its fixed gain parameters of the PD controller. 
Also note that our framework shows more accurate articulations on the in the-wild-sequence (see Fig.~\ref{fig:physcap_comparison}-(right)). 
In Fig.~\ref{fig:kinematic_comparison}, we compare our method with the state-of-the-art kinematic-based methods VNect \cite{VNect_SIGGRAPH2017} and VIBE \cite{kocabas2020vibe} on in-the-wild sequences. 
Only our method reconstructs smooth sequential 3D motions. 
The 3D motions by VNect and VIBE show sudden changes in joint positions which are observed as jitters in the  video. 

We next show the results of our approach with and without finetuning our network with 2D keypoints obtained on the sequences in the wild, see Fig.~\ref{fig:finetune} for the qualitative comparison. 
We use OpenPose \cite{openpose4} to obtain 2D keypoints, and the  networks are finetuned with the 2D reprojection loss. 
After the finetuning, our framework shows better overlay and visually more accurate 3D motions compared to the networks trained with the 3D benchmark datasets only (Human 3.6M, MPI-INF-3DHP and DeepCap).

\section{Conclusions}\label{sec:conclusions} 
We introduced a new fully-neural approach for 3D human motion capture from monocular RGB videos with hard physics-based constraints which runs at interactive framerates and achieves state-of-the-art results on multiple metrics. 
Our neural physical model allows learning motion priors and the associated physical properties, as well as gain values of the neural PD  controller from data. 
Thanks to the custom neural layer, which expresses hard physics-based constraints, our architecture is fully-differentiable. In addition, it can be trained jointly on several datasets thanks to the new form of input canonicalisation. 
Our experiments demonstrate that compared to  PhysCap---a recent method with physics-based boundary conditions---our physionical approach captures significantly faster motions, while being more accurate % 
in terms of various 3D reconstruction metrics. 
Thanks to the full differentiability, the proposed method can be finetuned on datasets with 2D annotations only, which improves the reconstruction fidelity on in-the-wild footages. 
These properties make it well suitable for direct virtual character animation from monocular videos, without requiring any further post-processing of the estimated global 3D poses. 

We believe that the proposed method opens up multiple directions for future research. 
Our architecture can be classified as a 2D keypoint lifting approach, which has both advantages (\textit{e.g.}, the possibility of 2D keypoint normalisation, on the one hand) and downsides (\textit{e.g.}, reliance on the accuracy of 2D keypoint detectors, on the other). 
Next, our results naturally lead to the question of what is the most effective way to integrate physics-based boundary conditions in neural architectures, and how 
the proposed ideas can be applied to many related problem settings. 
  
\appendix
\section{Network Details}\label{sec:appendix_netdetails}
 
\begin{figure}[b] 
\centering 
\includegraphics[width=\linewidth]{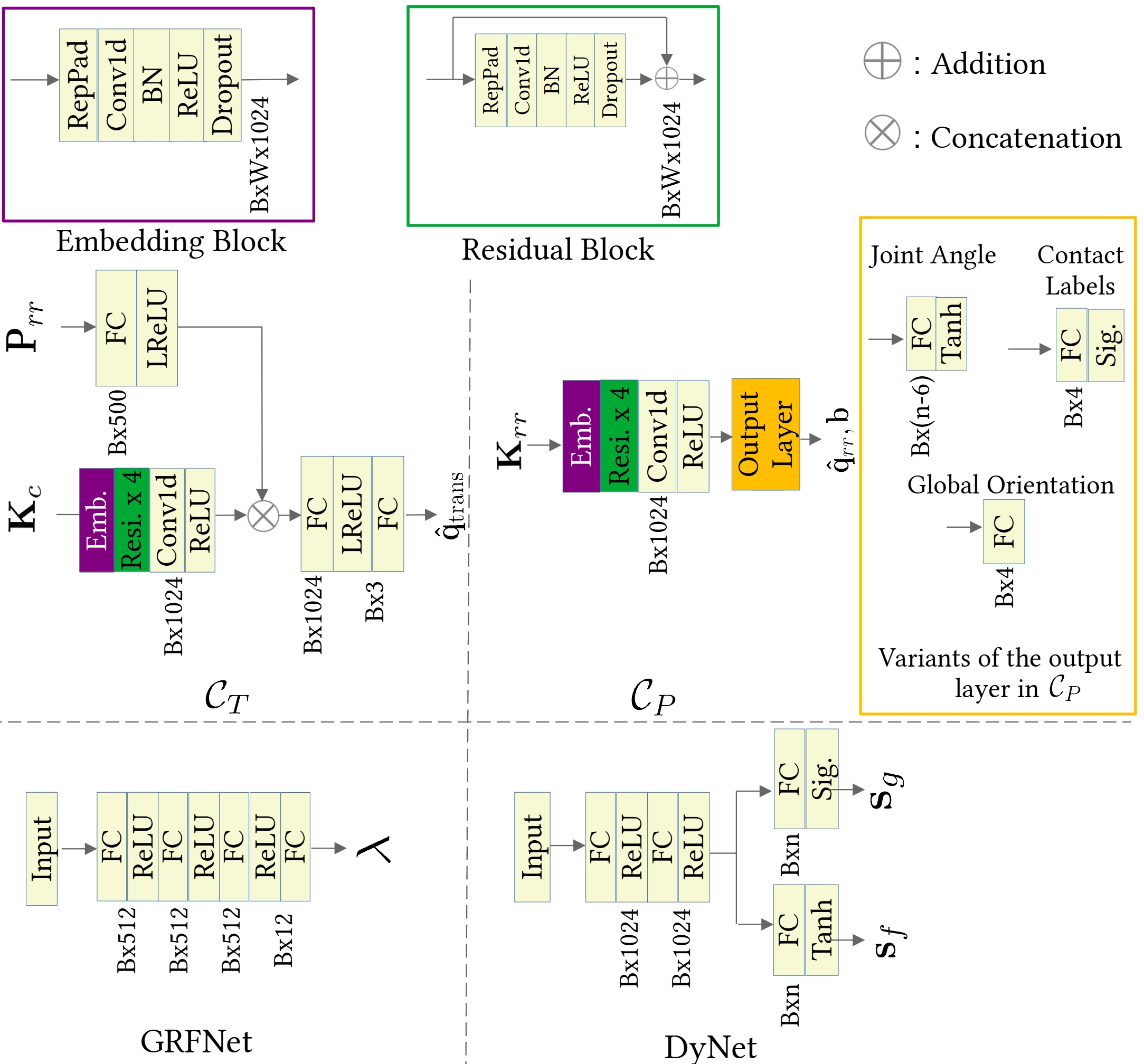} 
\caption{Schematic visualisations of the network details. ``Emb.'' and ``Resi.'' stand for the embedding block (purple box) and residual block (green box), respectively. ``BN'', ``RepPad'', ``FC'', ``Sig.'' and ``Conv1D'' represent batch normalisation, replication padding, fully-connected layer, sigmoid function and 1D convolution, respectively. The numbers next to the layers represent the output dimensionality. ``B'' and ``W'' represent the batch size and temporal window size, respectively.
} 
\label{fig:netarch} 
\end{figure}

We schematically visualise the network details in Fig.~\ref{fig:netarch}. 
Our implementations of $\mathcal{C}_{T}$ and $\mathcal{C}_{P}$ are based on  \cite{zou2020reducing} and composed of 1D convolutional layers with residual blocks. 
We use the replication padding layer of size $1$ for the embedding block and size $4$ for the residual block. 
The kernel size of the 1D convolutional layer for the embedding and residual blocks are $3$ and $5$, respectively. 
For the 1D convolution in the residual blocks, we use the dilation of size $2$.
For $\mathcal{C}_{T}$---although it is possible to estimate $\mathbf{\hat{q}}_{rr}$ and $\mathbf{b}$ with a single neural network---we observed that estimating the global rotation, joint angles and contact labels with three different networks shows higher accuracy.
Therefore, $\mathcal{C}_{P}$ consists of three replicated networks with the difference in the output layer, see Fig.~\ref{fig:netarch} for the details. For GRFNet and DyNet, all the inputs are concatenated to one vector and fed to the networks. 
We can estimate $\mathbf{k}_{p}$ and  $\alpha$ directly by DyNet, however, similar to  \cite{chentanez2018physics}, we obtain $\mathbf{s}_{g}$ and $\mathbf{s}_{f}$ 
($0<\mathbf{s}_{g}<1 $ and $-1\leq \mathbf{s}_{f}\leq 1$) using sigmoid and $\operatorname{tanh}$ functions, and compute $\mathbf{k}_{p} = 2\mathbf{s}_{g}\mathbf{k}^{\text{ini}}_{p}$ and $\alpha = \gamma\mathbf{s}_{f}$; $\mathbf{k}^{\text{ini}}_{p}$ denotes the initial gain parameters which are determined following \cite{shimada2020physcap}, and $\gamma$ is the coefficient which is determined empirically. 
Note that we use the fixed $\mathbf{k}^{\text{ini}}_{p}$ and $\gamma = 10$ values through all the experiments. 
We observed that this formulation leads to an improved  stability and faster convergence of the network training than directly estimating $\mathbf{k}_{p}$ and $\alpha$, since the network outputs are always within the normalised range. 

\begin{acks}
All data captures and evaluations were performed
at MPII by MPII. The authors from MPII were supported
by the ERC Consolidator Grant 4DRepLy (770784). We also acknowledge support from Valeo.
\end{acks} 
\bibliographystyle{ACM-Reference-Format}
\bibliography{sample-bibliography} 
\end{document}